\documentclass[letterpaper, 10 pt, conference]{ieeeconf}  % Comment this line out if you need a4paper

\IEEEoverridecommandlockouts                              % This command is only needed if 
\makeatletter
\def\endthebibliography{%
	\def\@noitemerr{\@latex@warning{Empty `thebibliography' environment}}%
	\endlist
}
\makeatother

\usepackage{graphics} % for pdf, bitmapped graphics files
\usepackage{epsfig} % for postscript graphics files
\usepackage{bbding}
\usepackage{times} % assumes new font selection scheme installed
\usepackage{url}
\usepackage{siunitx}
\usepackage{verbatim}
\usepackage{flushend}

\usepackage{booktabs}
\usepackage{multirow}
\usepackage{upgreek}
\usepackage{color}
\usepackage{amssymb}
\usepackage{amsmath}
\usepackage{graphicx}

\usepackage[ruled,vlined]{algorithm2e} % package for algorithm
\title{\LARGE \bf  Fiducial Tag Localization on a 3D LiDAR Prior Map }
\author{Yibo Liu, Jinjun Shan, Hunter Schofield% <-this % stops a space
	\thanks{This work was supported in part by NSERC Alliance Program under Grant ALLRP 555847-20, and in part by Mitacs Accelerate Program under Grant IT26108.}
	\thanks{The authors are with Department of Earth and Space Science and Engineering, York University, Toronto, Ontario M3J 1P3, Canada 
		{\tt\footnotesize \{yorklyb,jjshan,hunterls\}@yorku.ca}}%
}

\begin{document}
	
	\maketitle
	\thispagestyle{empty}
	\pagestyle{empty}

	%%%%%%%%%%%%%%%%%%%%%%%%%%%%%%%%%%%%%%%%%%%%%%%%%%%%%%%%%%%%%%%%%%%%%%%%%%%%%%%%
	\begin{abstract}
 % The LiDAR fiducial marker, akin to the well-known AprilTag used in camera applications, serves as a convenient resource to impart artificial features to the LiDAR sensor, facilitating robotics applications. Unfortunately, current LiDAR fiducial marker detection methods are limited to occlusion-free point clouds. In this work, we present a novel approach for occlusion-resistant LiDAR fiducial marker detection. We first extract 3D points potentially corresponding to the markers, leveraging the 3D intensity gradients. Afterward, we analyze the 3D spatial distribution of the extracted points through clustering. Subsequently, we determine the potential marker locations by examining the geometric characteristics of these clusters. We then successively transfer the 3D points that fall within the candidate locations from the raw point cloud onto a designed intermediate plane. Finally, using the intermediate plane, we validate each location for the presence of a fiducial marker and compute the marker's pose if found. We conduct both qualitative and quantitative experiments to demonstrate that our approach is the first LiDAR fiducial marker detection method applicable to point clouds with occlusion while achieving better accuracy.
The LiDAR fiducial tag, akin to the well-known AprilTag used in camera applications, serves as a convenient resource to impart artificial features to the LiDAR sensor, facilitating robotics applications. Unfortunately, the existing LiDAR fiducial tag localization methods do not apply to 3D LiDAR maps while resolving this problem is beneficial to LiDAR-based relocalization and navigation. In this paper, we develop a novel approach to directly localize fiducial tags on a 3D LiDAR prior map, returning the tag poses (labeled by ID number) and vertex locations (labeled by index) w.r.t. the global coordinate system of the map. In particular, considering that fiducial tags are thin sheet objects indistinguishable from the attached planes, we design a new pipeline that gradually analyzes the 3D point cloud of the map from the intensity and geometry perspectives, extracting potential tag-containing point clusters. Then, we introduce an intermediate-plane-based method to further check if each potential cluster has a tag and compute the vertex locations and tag pose if found.  We conduct both qualitative and quantitative experiments to demonstrate that our approach is the first method applicable to localize tags on a 3D LiDAR map while achieving better accuracy compared to previous methods. The open-source implementation of this work is available at: https://github.com/York-SDCNLab/Marker-Detection-General.

% We first extract 3D points potentially corresponding to the tags, leveraging the 3D intensity gradients. Afterward, we analyze the 3D spatial distribution of the extracted points through clustering. Subsequently, we determine the potential tag locations by examining the geometric characteristics of these clusters. We then successively transfer the 3D points that fall within the candidate locations from the raw point cloud onto a designed intermediate plane. Finally, using the intermediate plane, we validate each location for the presence of a fiducial tag and compute the tag's pose if found.  We conduct both qualitative and quantitative experiments to demonstrate that our approach is the first method applicable to localize tags on a 3D LiDAR map while achieving better accuracy compared to previous methods. The open-source implementation of this work is available at: https://github.com/York-SDCNLab/Marker-Detection-General.

	\end{abstract}

	 %This kind of point cloud could be, but is not limited to, the 3D map built from Simultaneous Localization and Mapping (SLAM).
	%%%%%%%%%%%%%%%%%%%%%%%%%%%%%%%%%%%%%%%%%%%%%%%%%%%%%%%%%%%%%%%%%%%%%%%%%%%%%%%%
	\section{INTRODUCTION} \label{intro}
Delivery based on unmanned systems, such as drones, is gaining popularity. A common solution to provide accurate localization for unmanned systems is placing fiducial tags, such as AprilTag \cite{ap3} and ArUco \cite{aruco}, in the environment. However, the visual fiducial tags are sensitive to ambient light, and thus cannot function in a dark or an overexposed environment. On the other hand, with the development of Light Detection and Ranging (LiDAR) sensors, the cost of LiDAR has been decreasing. Thus, more and more unmanned systems are being equipped with LiDARs \cite{loam,sdk}. The LiDAR sensor is not affected by illumination \cite{lt}. Recently, there have been works developing fiducial tags for the LiDAR sensor \cite{IFM,lt,a4}. If unmanned systems can navigate using LiDAR fiducial tags, they would be applicable in darkness or overexposure and robust to changes in ambient light.
\begin{figure}[thpb]
		\centering
		\includegraphics[width=3.3in]{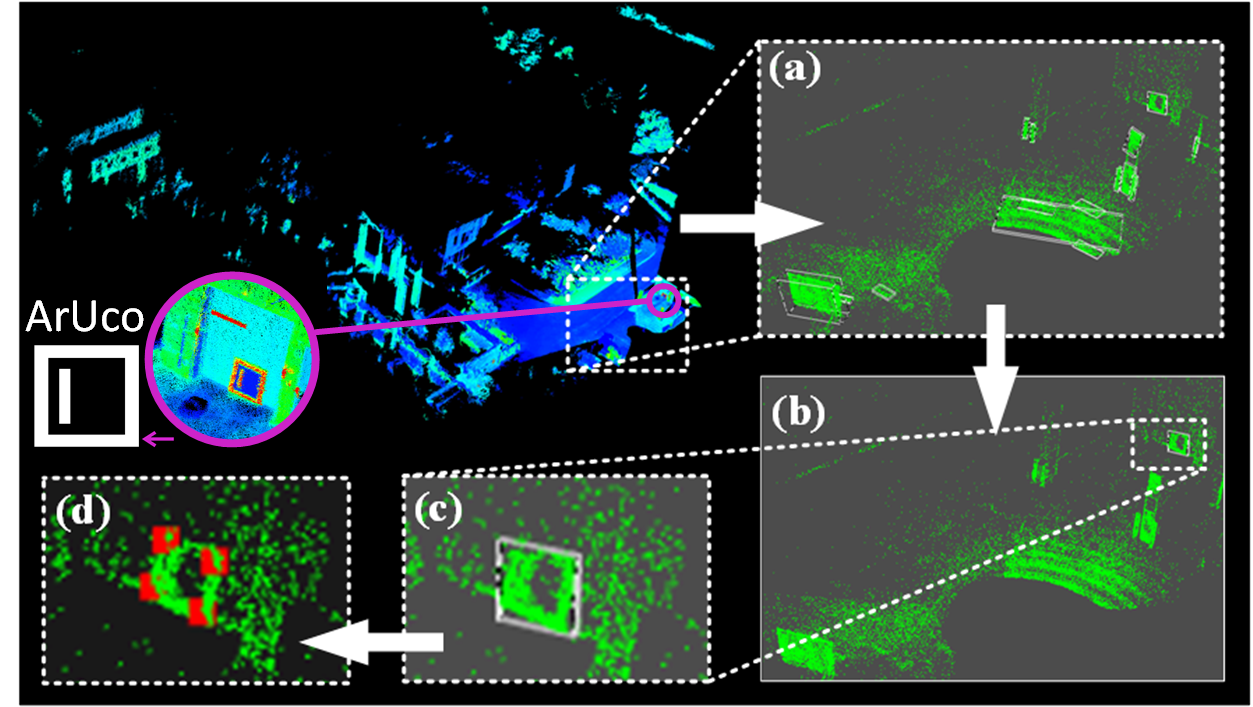}
		\caption{An illustration of the proposed method, which jointly analyzes the point cloud of the 3D map from both intensity and geometry perspectives, returning the tag poses and vertex locations w.r.t. the global coordinate system of the map. (a): After sections \ref{ic} and \ref{clu}. (b): After section \ref{clu2}. As seen, all the clusters but the one belonging to the ArUco tag \cite{aruco} are filtered out. (c): The zoomed view of the preserved oriented bounding box. (d): A zoomed view of the detected 3D fiducials, which are rendered in red.}
	% \caption{  An illustration of the tag localization process on the 3D LiDAR map shown in Fig. \ref{mov}. (a): After sections \ref{ic} and \ref{clu}. (b): After section \ref{clu2}. As seen, all the clusters but the one belonging to the tag are filtered out. (c): The zoomed view of the preserved OBB. (d): A zoomed view of the detected 3D fiducials, which are rendered in red.}

		\label{mov}
	\end{figure} \par
\par
The prerequisite for building a tag-based navigation system is to localize the tags in the environment, typically represented by a 3D map \cite{map3}. Unfortunately, in previous studies, the poses of tags on a prior map are usually measured manually, leading to both inaccurate localization and much human labor. \cite{map3} proposes to automatically localize fiducial tags on a 3D map by utilizing tags as landmarks during the mapping procedure. However, this does not apply to LiDAR-based applications because LiDAR fiducial tags \cite{IFM,lt,a4}, unlike the mature visual fiducial tags utilized in \cite{map3}, just emerged in recent years and have not yet been applied to LiDAR-based Simultaneous Localization and Mapping (SLAM) \cite{sdk,loam}. Therefore, unless a bottom-level modification is applied to the frontend of LiDAR-based SLAM, it is not straightforward to follow \cite{map3} to employ fiducial tags as landmarks in LiDAR-based SLAM \cite{sdk,loam}.
\par
% Map-based localization methods \cite{map1,map2,map3} are beneficial to downstream tasks such as autonomous navigation, and thus, they have been gaining popularity in recent years. In particular, these methods can precisely localize the robot/unmanned vehicle in a 3D prior map by place recognition, which provides positional information for navigation \cite{map4}. Nevertheless, for environments lacking natural features, these localization methods become unstable. Consequently, fiducial tags, such as AprilTag \cite{ap3}, ArUco \cite{aruco}, and CCTag \cite{cctag}, are crucial in these cases as they provide artificial features (a.k.a. fiducials). By placing fiducial tags in the environment and integrating them into the localization method, it is feasible to build up a reliable navigation system in a featureless environment \cite{map5,map6,yibo}. \par

% \vspace{-0.2cm}
% Determining the poses of fiducial tags on a 3D prior map is important since the tag-based localization accuracy directly has an impact on the navigation. 
\begin{table*}[htbp]
\caption{Comparision with existing methods}
\begin{center}
% \centering

\begin{tabular}{c|c|c|c|c}
\hline\hline
%&\multicolumn{3}{|c|}{\textbf{Table Column Head}} \\
Methods & Handles 3D Map&Input & Tag Type &Core Algorithm   \\
\cline{1-5} 
LiDARTag \cite{lt} & No & single-view scan &Extra 3D object & Geometry Clustering  \\ \hline
A4LiDARTag \cite{a4}& No & single-view scan & Extra 3D object &Range Image   \\ \hline
IFM \cite{IFM}&No & single-view scan &Thin paper &Intensity Image   \\ \hline
Ours & Yes & 3D map or single-view scan &Thin paper &Joint analysis of intensity and geometry\\  \hline\hline

\end{tabular}
\label{tab0}
\end{center}
\end{table*}
An alternative solution is to adopt the conventional SLAM method \cite{sdk,loam} to build up the map first and then localize the fiducial tags on the map. However, the existing LiDAR tag detection/localization methods \cite{IFM,lt,a4} do not apply to a 3D LiDAR map because they are designed to deal with the LiDAR scan obtained from a single view (\textit{i.e.} the LiDAR is stationary).
To this end, we develop a novel method, as depicted in Fig. \ref{mov}, to directly localize LiDAR fiducial tags on a 3D environmental map. Given a 3D LiDAR map, the proposed method automatically recognizes each tag (labeled by a unique ID number) and returns its 6-DOF pose together with its vertice positions (labeled by vertice index). 
%
% With our method, it is feasible to construct a LiDAR-based localization system as shown in Fig. \ref{mov}. \par
%
Considering that the fiducial tags in this work refer to thin sheets of paper that are indistinguishable from the planes to which they are attached, we design a new pipeline to jointly analyze the point cloud of the map from intensity and geometry perspectives. With the proposed pipeline, we extract the potential tag-containing point clusters from the point cloud. Then, we introduce an intermediate-plane-based method to check whether each cluster contains a tag and compute its pose and vertex locations if found.
We carry out experiments on both synthetic and real-world scenes against existing methods to demonstrate that our method is the first algorithm capable of directly localizing fiducial tags on a 3D map while achieving higher accuracy. In addition, we release the open-source implementation at https://github.com/York-SDCNLab/Marker-Detection-General. \par

The main \textbf{contributions} of this work are as follows:
 \begin{itemize}
% \vspace{-0.2cm}
\item We propose a novel method to localize fiducial tags on a 3D LiDAR map, which is a problem that cannot be resolved by the existing approaches \cite{IFM,lt,a4}. A comparison of our method and the existing methods is presented in Table \ref{tab0}.
% \vspace{-0.2cm}
\item We design a new pipeline to analyze a 3D point cloud from the intensity and geometry perspectives jointly, 
considering that fiducial tags are planar objects with high-intensity contrast and are non-distinguishable from the plane to which they are attached. 
This differs from conventional 3D object detection methods that rely merely on 3D geometric features and can only detect spatially distinguishable objects.
% \vspace{-0.2cm} \par
\end{itemize} \par

\section{Methodology} 
\subsection{Preliminaries} \label{pre}
In this work, the fiducial tags refer to sheets of thin paper. As shown in Table \ref{tab0}, among the existing methods, only IFM \cite{IFM} is applicable to this type of tag. However, IFM is incapable of dealing with a 3D map. In this section, we introduce preliminaries regarding spherical projection (Eq. (\ref{pro})) to further explain the limitations of IFM and illustrate the necessity of developing a new pipeline to analyze the point cloud.

\par
IFM first projects the 3D points of the raw point cloud into a 2D image plane to construct the intensity image and then adopts the 2D marker detector to find 2D fiducials on it. Finally, the detected 2D fiducials are projected back into 3D space through the reverse spherical projection. In particular, the intensity image is generated by transforming each 3D point in the raw point cloud into a 2D pixel on an image plane:
\begin{equation}	
   u = \lceil \frac{\theta}{\Theta_{a}}\rfloor + u_{o},\; v = \lceil \frac{\phi}{\Theta_{i}}\rfloor + v_{o},
\label{pro}
\end{equation}
where $[u,v]^{T}$ are the image coordinates of the projected pixel. $[\theta,\phi,r]^{T}$ are the spherical coordinates of the 3D point. $\Theta_{a}$ and $\Theta_{i}$ are the angular resolutions in $u$ (azimuth) and $v$ (inclination) directions, respectively. $u_{o}$ and $v_{o}$ are the offsets. Apparently, for any two points sharing the same $[\theta,\phi]^{T}$, they will overlap for the same pixel. This indicates that occlusion ruins the intensity image due to pixels overlapping. \par
\begin{figure}[thpb]
		\centering
		\includegraphics[width=2.0in]{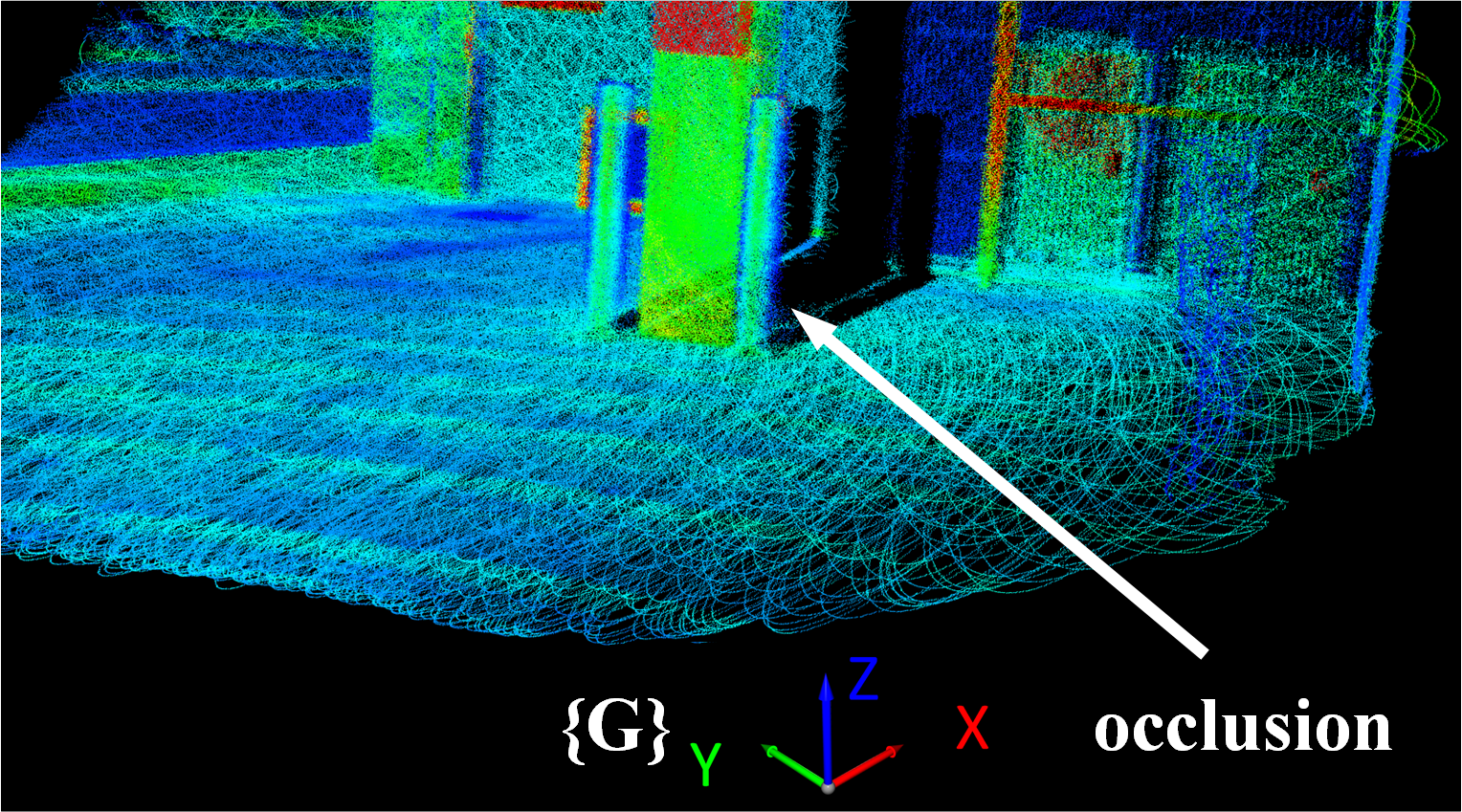}
		\caption{The 3D map shown in Fig. \ref{mov} observed from the origin of the global coordinate system. The occlusion is not an artificial challenge, and a detailed explanation is provided in Remark 1.}
		\label{occ}
	\end{figure}

Now, we explain why IFM \cite{IFM} or the spherical projection (Eq. (\ref{pro})) is not applicable to a 3D LiDAR environmental map as shown in Fig. \ref{mov}. First, if we observe the 3D map from a bird perspective as displayed in Fig. \ref{mov}, there is no occlusion between the fiducial tag and us. However, this does not indicate the spherical projection is applicable because the perspective shown in Fig. \ref{mov} is manually tuned for the display purpose. As shown in Fig. \ref{occ}, if we observe the map from the origin of the global coordinate system of the map, the tag is invisible due to occlusion. Thus, it is desired to develop a new solution to resolve this problem.  
 \par
 
\noindent\textbf{Remark 1.} One might think that we can shift the viewpoint to obtain a better perspective with no occlusion and then apply IFM \cite{IFM} to detect the tag. 
%
% However, for a constructed 3D LiDAR map already saved as a point cloud file, we only know the global coordinate system, namely, one perspective. Thus, we have to either manually adjust the viewpoint, which is human effort, or design an algorithm to automatically identify such optimal perspectives. 
%
However, unlike humans, who can judge the quality of a perspective intuitively, robots or computers must iteratively assess potential perspectives, which requires an unimaginably high computational cost.  \par
\noindent\textbf{Remark 2.} 
Given the explicit visibility of the tag in Fig. \ref{mov}, one might think that the fiducial tag is scanned without occlusion. Therefore, by applying IFM when there is no occlusion during SLAM, the fiducial tag can be localized. We do not deny the feasibility of this approach. However, fiducial tag detection by LiDAR is not yet integrated into LiDAR-based SLAM, which is not trivial as it involves bottom-level modification to the frontend of SLAM. Thus, we opt to first construct the map and then localize the tags on the built map.\par
  
	\begin{figure}[thpb]
		\centering
		\includegraphics[height=1.0in]{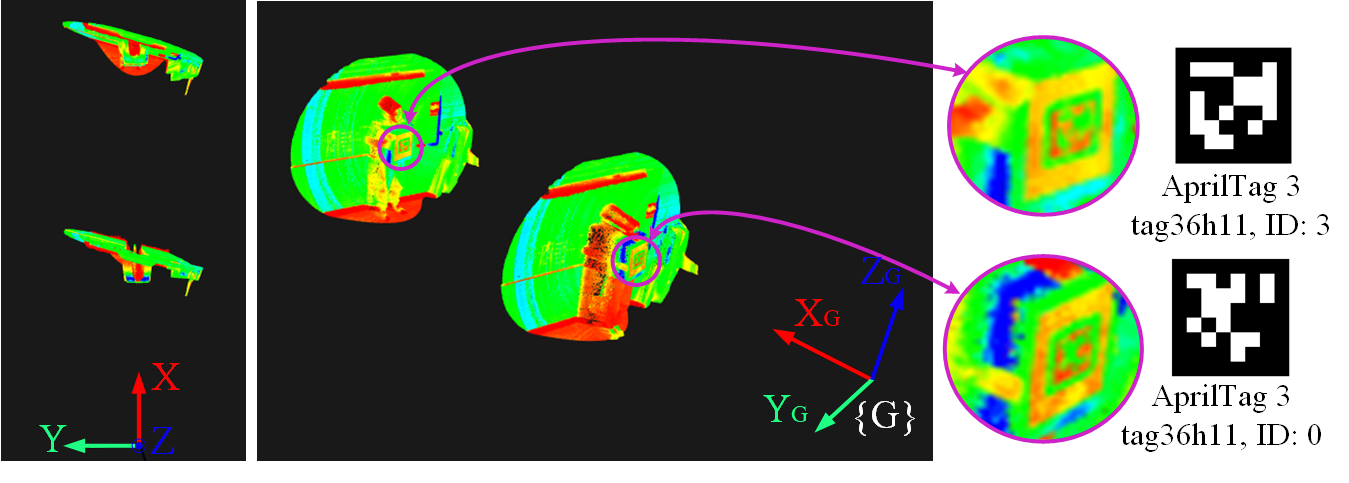}
		\caption{The example used to explain the design purpose and result of each step.
 The two presenters are holding two different AprilTags \cite{ap3}. As illustrated in the top view, observing along the X-axis of the global coordinate system, the back subpoint cloud is totally blocked by the front one. Thus, Eq. (\ref{pro}) cannot function in this case.}
		\label{sub}
	\end{figure} \par
\subsection{Downsampling Based on 3D Intensity Gradients} \label{ic} 
We utilize a synthesis point cloud (see Fig. \ref{sub}) with a simple scene to explicitly illustrate the design purpose and results of each operation in our pipeline. This is not a 3D LiDAR map, but it possesses the most important feature of a 3D LiDAR map in this research: the spherical projection (Eq. (\ref{pro})) is not applicable.

     	\begin{figure}[thpb]
		\centering
		\includegraphics[width=3.3in]{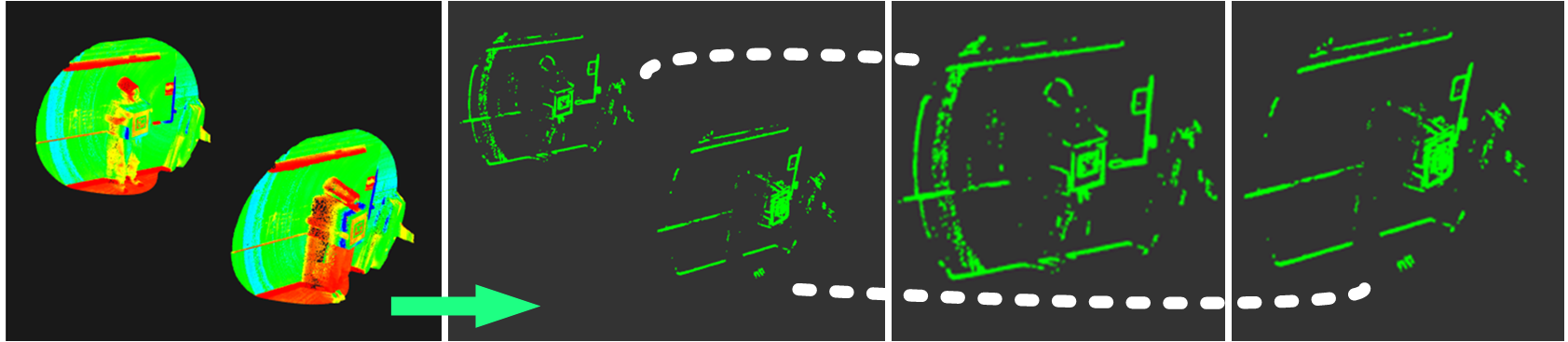}
		\caption{The effect of applying downsampling from the intensity perspective. As seen, the majority of unnecessary points are filtered out at this step.}
		\label{step1}
	\end{figure} \par
    	
As declared above, the fiducial tags in this work refer to sheets of thin paper that are not spatially distinguishable from the attached planes. Thus, it is infeasible to adopt previous geometric features-based 3D object detection methods \cite{single, rangenet,voxelnext} to find the fiducial markers. Despite this, a fiducial tag is composed of a black-and-white pattern and, as a result, presents as a high-intensity contrast object in the view of a LiDAR (see the zoomed views of Fig. \ref{mov} and Fig. \ref{sub}). This indicates that we can first analyze the point cloud from the intensity perspective. \par
In particular, we conduct downsampling on the raw point cloud based on the 3D intensity gradients. We take the intensity as a function $I(\mathbf{x})$ of the 3D coordinates $\mathbf{x} = [x,y,z]^{T}$. Suppose that the given 3D point/location is $\textbf{p}_{0}$, we define the point set composed of the neighbouring $n$ points around $\textbf{p}_{0}$ as $\mathcal{P}_{I}=\{\textbf{p}_{1}, \textbf{p}_{2},\cdots,\textbf{p}_{n}\}$. In practice, we use a linear model, $\hat{I}(\mathbf{x})$, to approximate $I(\mathbf{x})$:
\begin{equation}
\hat{I}(\mathbf{x}) = \mathbf{A}^{T}\mathbf{x} + b,
\end{equation}
where $\mathbf{A} \in \mathbb{R}^{3 \times 1}$ and $b = \bar I$.  $\bar I \in \mathbb{R}$ is the mean of the intensity values of the points in $\mathcal{P}_{I}$. Since the intensity values of points in $\mathcal{P}_{I}$ are known, we can estimate $\mathbf{A}$ by resolving the following model fit (least square) problem:
\begin{equation}	
\underset{\mathbf{A}}{\arg \min } \sum_{i=1}^{n}\left\|\hat{I}(\mathbf{x}_{i})-I(\mathbf{x}_{i})\right\|^{2}\label{least}.
\end{equation} \par
Once $\hat{I}(\mathbf{x})$ is obtained, we can compute the 3D intensity gradients $\nabla I \in \mathbb{R}^{3 \times 1}$ at $\textbf{p}_{0}$. %
% \begin{equation}
% \nabla I=\left[\begin{array}{lll}
% \frac{\partial I}{\partial x} & \frac{\partial I}{\partial y} & \frac{\partial I}{\partial z}
% \end{array}\right]^{T}
% \end{equation}
% where $\frac{\partial I}{\partial x}$, $\frac{\partial I}{\partial y}$ and $\frac{\partial I}{\partial z}$ describe the intensity changes in the x, y, and z directions, respectively.
% \par
The direction of $\nabla I$  of a given point indicates the direction where the intensity has the fastest decline and the norm, $|\nabla I|$, implies the rate of descent. 
Inspired by LOAM \cite{loam}, which selects a point as a feature if its geometric curvature is larger than a threshold, we preserve the 3D point if its $|\nabla I|$ is larger than a threshold in the downsampling procedure. The result following downsampling execution is depicted in Fig. \ref{step1}. \par
\noindent\textbf{Remark 3.} One might argue that since Fig. \ref{sub} is a stitch of two point clouds without occlusion, it appears that we can localize tags on them separately using IFM \cite{IFM}. Consequently, the analysis in this section and the following sections seems to focus on an artificial problem. Please note that we only use Fig. \ref{sub} to explain the results and design purposes due to its simplicity. Our ultimate objective is to localize the fiducial tags on a 3D LiDAR map, which represents a larger and more complex scene. For 3D LiDAR maps, Remarks 1 and 2 elucidate the challenges and our motivations.
\subsection{Spatial Distribution Analysis of Downsampling Result} \label{clu}
\begin{figure}[thpb]
		\centering
		\includegraphics[width=1.8in]{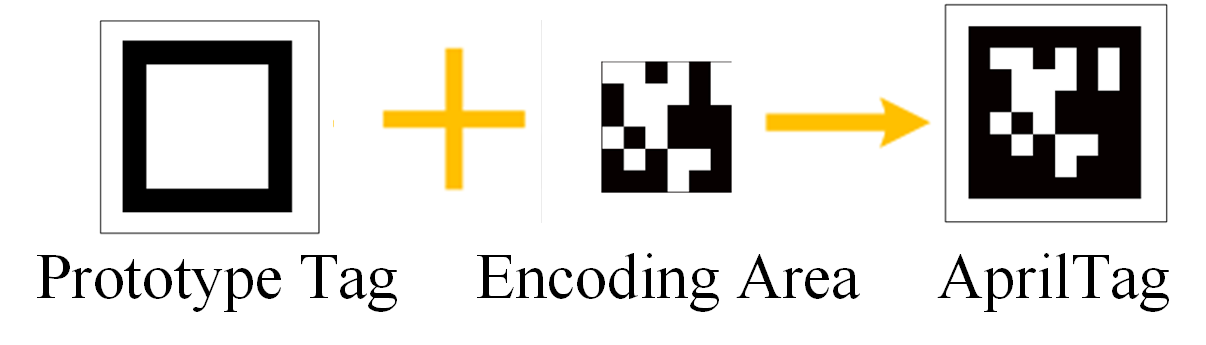}
		\caption{ A diagram to illustrate the design of a typical square fiducial tag \cite{ap3}. A square fiducial tag is a combination of the prototype tag (a black frame inside a white frame) and the encoding area.}
		\label{td}
	\end{figure}

The downsampling preserves the points belonging to the outlines of all objects with high intensity-contrast. In this section, we analyze the point cloud from the geometric perspective.
% are preserved after the feature extraction. To further analyze if a feature belongs to the LiDAR fiducial marker or to a naturally occurring object, we conduct clustering on the point cloud of the features. 
The foundation of doing so is the fact that the points belonging to the fiducial tags will be isolated from those of the other objects after the downsampling (see the zoomed view of  Fig. \ref{step1}). This is due to the design of the tag's pattern as shown in Fig. \ref{td}. 
 \par
\begin{figure}[thpb]
		\centering
		\includegraphics[width=3.3in]{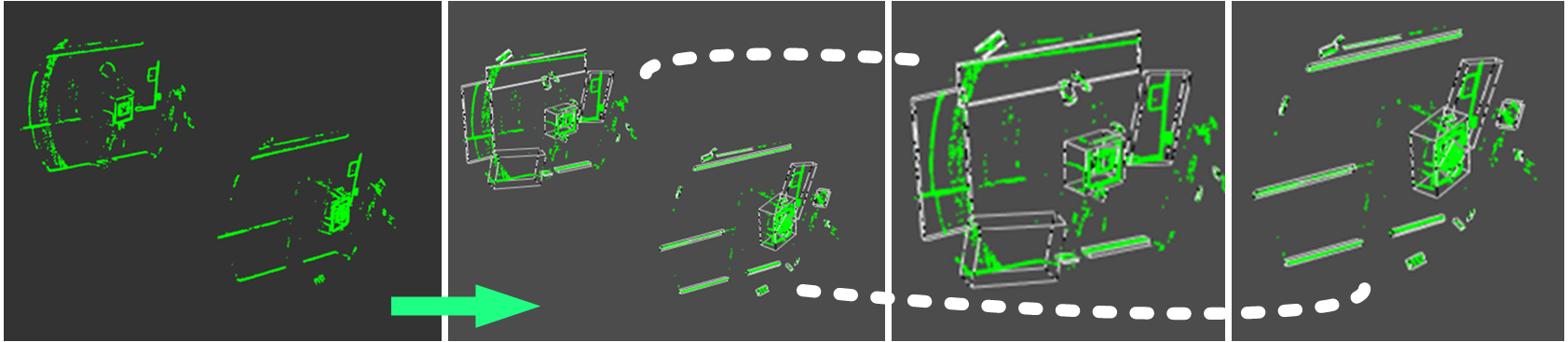}
		\caption{The effect of clustering on the downsampling result.}
		\label{step2}
	\end{figure}

% In the visual fiducial tag system, such as AprilTag \cite{ap3}, the aim of adopting the prototype tag (the black-white protector frame) is to isolate the tag from the environment and to construct a square shape that can be easily recognized by the visual quad detector. Despite the fact that the design of the prototype marker does not consider any factors related to LiDAR, we found that the prototype tag also brings an important benefit to this LiDAR application: 
The white regions of the prototype tag naturally have higher intensity values than the black regions, and thus, the prototype tag after the downsampling is rendered as a square double-ring that isolates the points inside the coding area from the environment. We utilize this benefit to develop the follow-up pipeline. The isolation makes the points belonging to the tags spatially distinguishable from those of the other objects. Thus, we employ the method introduced in \cite{rusu} to further segment the downsampling result into clusters. Each cluster is represented by an oriented bounding box (OBB) as shown in Fig. \ref{step2}. \par
     	 \par
 \vspace{-0.2cm}
\subsection{Filtering Out Unwanted Clusters} \label{clu2}
As depicted in Fig. \ref{step2}, there are many OBBs after clustering. In this section, we analyze the geometric characteristic of each OBB to verify if it has the potential to be a fiducial tag. Specifically, the bounding box of a cluster needs to satisfy two criteria to be recognized as a candidate that could contain a tag. \par
\noindent\textbf{Criterion 1.} The first criterion is subject to the tag size: 
\begin{equation}	
		\sqrt{2a^{2}+\delta^{2}}\leq L_{OBB} \leq \sqrt{4a^{2}+\delta^{2}},
		\label{first}
	\end{equation} 
where $L_{OBB}=\sqrt{l^{2}+w^{2}+h^{2}}$ is the cuboid diagonal of the OBB with $l$, $w$, and $h$ ($h \leq \delta$) being the length, width, and height, respectively. $a$ denotes the side length of the tag. $\delta$ is the trifling thickness of the tag. \par
\begin{figure}[thpb]
		\centering
		\includegraphics[width=2.0in]{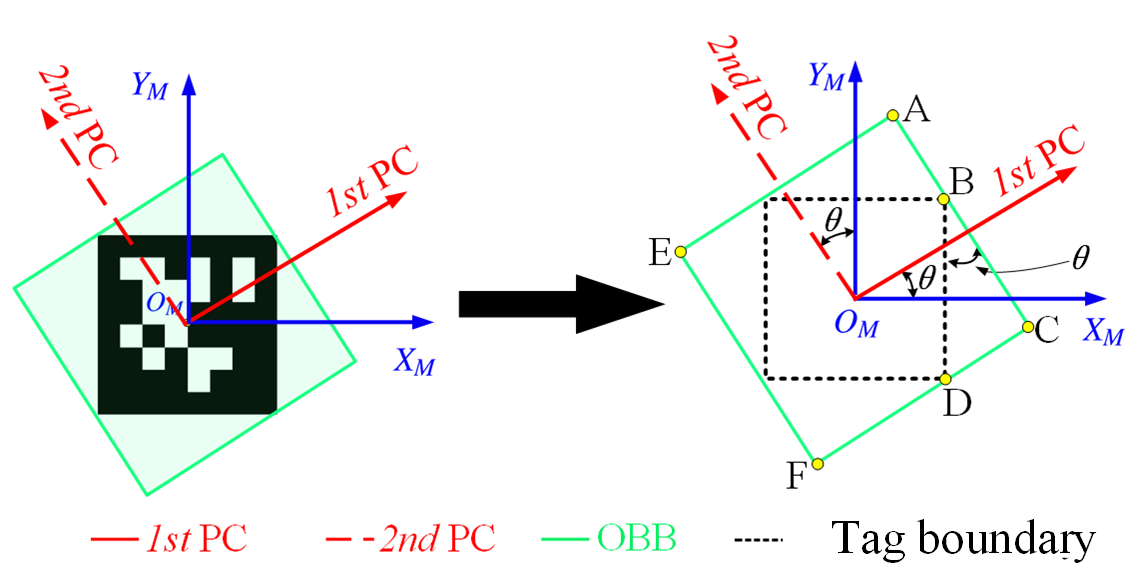}
		\caption{ A diagram of the possible OBB size for a given tag. $O_{M}-X_{M}-Y_{M}$ is the tag coordinate system. $\theta$ is the angle between the first primary component (PC) and the x-axis. On account of the fact that the first PC and second PC are perpendicular to each other \cite{bounding}, the angle between the second PC and the y-axis is $\theta$. In addition, $\angle DBC=\theta$.}
		\label{proof}
	\end{figure}

\noindent\textbf{Explanation of Criterion 1.} We neglect the trivial height for now and consider the possible size range of the OBB in 2D space. According to the simple geometry derivation given in the caption of Fig. \ref{proof}, we have $a (\mathrm{cos}\theta +\mathrm{sin}\theta) \in [a, \sqrt{2}a]$ considering that $\theta \in [0, 2\pi]$. Thus, the area of the 2D OBB, $S_{OBB}$, is in the following range:
	\begin{equation}	
		a^{2}\leq S_{OBB} \leq 2a^{2}.
		\label{range}
	\end{equation} \par
Now, returning to the 3D space, we also take into account the tag's thickness, $\delta$. The first criterion becomes the one shown in Eq. (\ref{first}).
	 \par
 % 	\begin{figure}[thpb]
	% 	\centering
	% 	\includegraphics[width=2.0in]{pcall}
	% 	\caption{A rough schematic of the smallest (a) and the largest (b) OBBs. (a):  \textit{1st} PC $\parallel$ $s_{1}$ and \textit{2nd} PC $\parallel$ $s_{2}$ where $s_{1}$ and $s_{2}$ are the horizontal side and vertical sides of the tag, respectively. (b):  \textit{1st} PC $\parallel$ $d_{1}$ and \textit{2nd} PC $\parallel$ $d_{2}$ where $d_{1}$ and $d_{2}$ being the two diagonals of the tag.}
	% 	\label{pcall}
	% \end{figure} \par

\noindent\textbf{Criterion 2.} The second criterion is shown in Eq. (\ref{second}): 
	\begin{equation}	
		 1/1.5 \leq l/w \leq 1.5.
		\label{second}
	\end{equation} \par

\begin{figure}[thpb]
		\centering
		\includegraphics[width=3.3in]{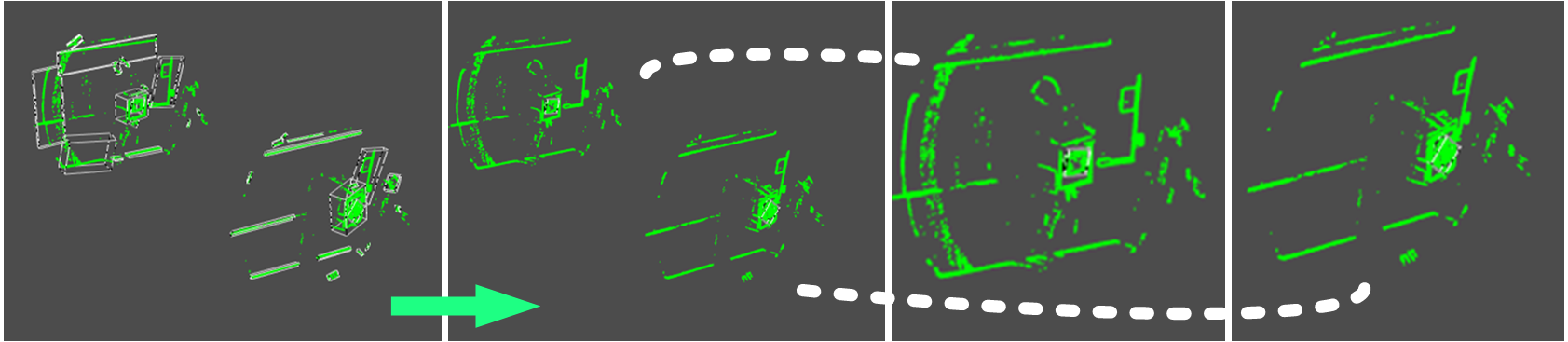}
		\caption{The effect of filtering out the unwanted clusters through the two criteria.}
		\label{step3}
	\end{figure}

\noindent\textbf{Explanation of Criterion 2.} This criterion is based on the fact that the shape of the tag is square and the OBB projection on the plane of length and width is also square. Ideally, we have $l \approx w$. Whereas in the real world, the tag cannot be perfectly scanned by LiDAR, and LiDAR also has ranging noise. As a result, the shape of the OBB on the length and width plane could be distorted. Hence, we make the requirement on $l \approx w$ less strict as shown in Eq. (\ref{second}). It is presented in Fig. \ref{step3} that the unwanted OBBs are filtered out after the operation introduced in this section. 
 	 \par
\subsection{Tag Detection via Intermediate Plane\label{ip}}
After analyzing the point cloud from the 3D intensity gradients and 3D geometric perspectives, we obtain the locations of objects with high intensity-contrast and shapes/sizes similar to the fiducial tag. 
We then inspect these locations (OBBs) in the raw point cloud. When extracting points falling into the OBBs, a buffer is adopted to extend the OBBs, preserving more regions around an OBB in case it does not completely cover the fiducial tag. This indicates that, for each OBB, the pose (position and orientation), $\mathbf{T}_{OBB}$ \footnote{$\mathbf{T}_{OBB}$ denotes the transmission from the global coordinate system $\{\mathbf{G}\}$ to the OBB frame. The OBB frame refers to a coordinate system whose X, Y, and Z axes are parallel to the length, width, and height of the bounding box. The origin of the OBB frame is located at the center of the bounding box. }, is kept while the size is enlarged by multiplying the length, width, and height with an amplification factor, $t_{b}$. The recommended value of $t_{b}$ is twice the tag's side length based on our experiment. \par

 	\begin{figure}[thpb]
		\centering
		\includegraphics[width=3.3in]{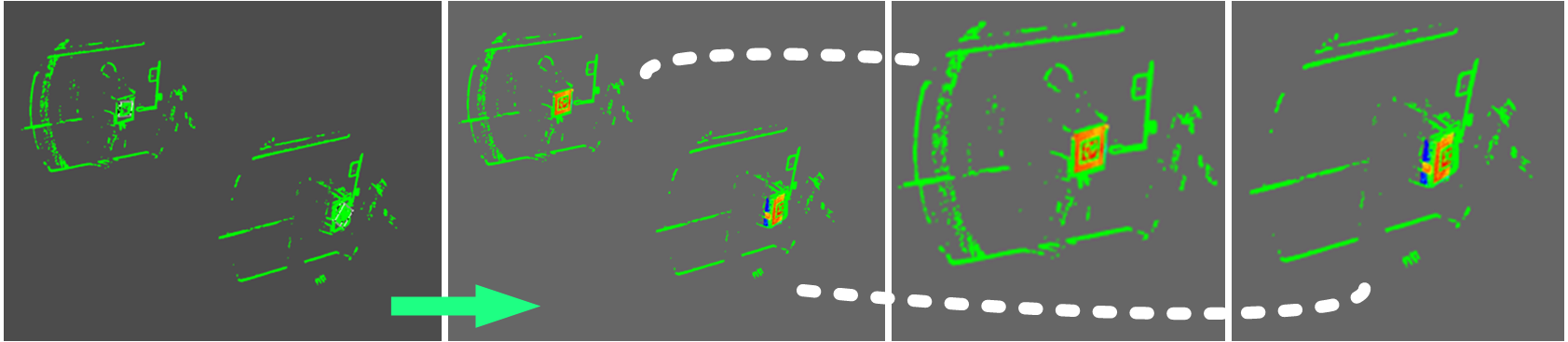}
		\caption{The result of extracting points falling into the preserved OBBs from the raw point cloud.}
		\label{step4}
	\end{figure} \par
\begin{figure}[thpb]
		\centering
		\includegraphics[width=3.0in]{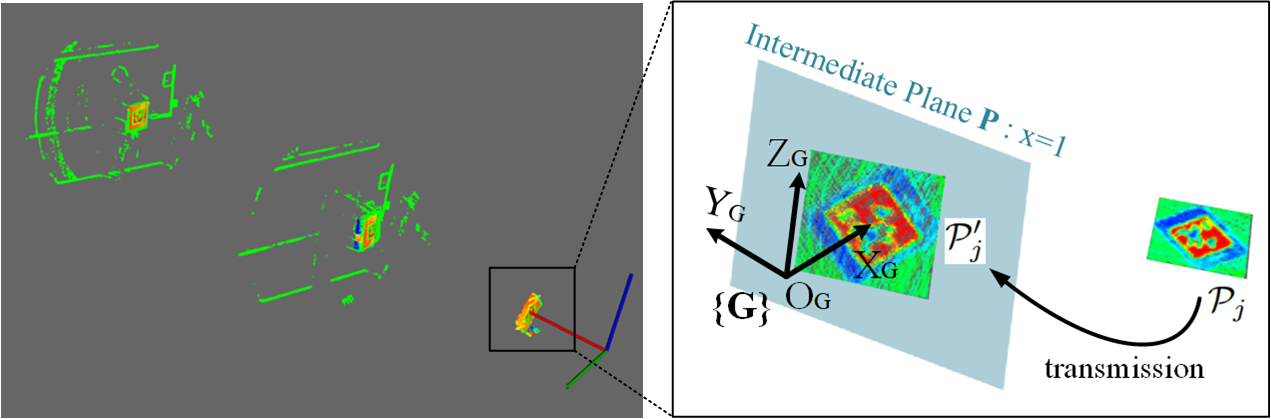}
		\caption{An illustration of the intermediate plane.}
		\label{inter}
	\end{figure}

\noindent\textbf{Motivation for adopting an intermediate plane.} As seen in Fig. \ref{step4}, although the points belonging to the candidate tags are extracted from the raw point cloud, unfortunately, we still cannot apply the spherical projection in this case since occlusion still exists when observing the point cloud from the origin (review Fig. \ref{sub} if needed). Despite this, we already obtained the pose of each OBB ($\mathbf{T}_{OBB}$) through the analysis in sections \ref{ic}, \ref{clu}, and \ref{clu2}. Now we can adjust the perspective observing these candidate tags using $\mathbf{T}_{OBB}$. To eliminate occlusion, we choose to fix the viewpoint at the origin and transfer the points extracted from each OBB one by one to an intermediate plane (denoted by $\mathbf{P}$), as shown in Fig. \ref{inter}.
	 \par
\begin{figure}[thpb]
		\centering
		\includegraphics[width=3.3in]{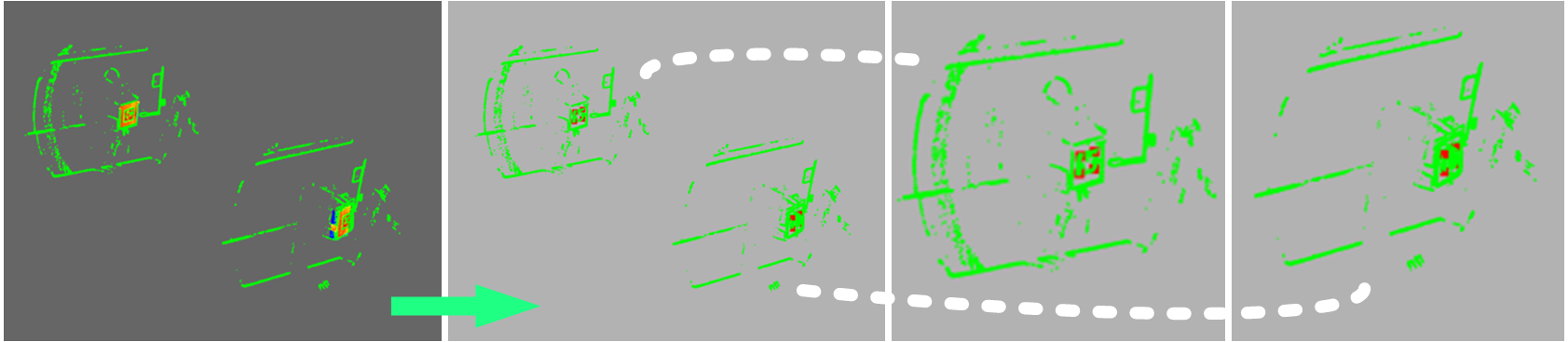}
		\caption{The fiducial tag localization result after applying the intermediate plane method. The vertices of the localized fiducial tag are rendered red.}
		\label{step5}
	\end{figure}

The detailed transmission process is as follows. Define the point set of the $j$th OBB as $\mathcal{P}_{j}$ and a point of $\mathcal{P}_{j}$ as $\textbf{p}_{j}\in \mathbb{R}^{3}$. $\textbf{p}_{j}$ is first transmitted to the origin of the global coordinate system $\{\mathbf{G}\}$ through the inverse of $\mathbf{T}_{OBB}$:
	\begin{equation}	
			\textbf{p}_{G}  = \mathbf{T}_{OBB}^{-1} \cdot \textbf{p}_{j}= \mathbf{R}^{-1} \cdot \textbf{p}_{j}-\mathbf{R}^{-1}\mathbf{t}, 
		\label{trans}
	\end{equation}
	where $\textbf{p}_{G}$ is the point transmitted to the origin of $\{\mathbf{G}\}$. $\mathbf{T}_{OBB} = [\mathbf{R}|\mathbf{t}]$ where $\mathbf{R}$ is the $3 \times 3$ orthogonal rotation matrix and $\mathbf{t}$ is the $3 \times 1$ translation vector. After transmitting all the points in $\mathcal{P}_{j}$ using Eq. (\ref{trans}), we obtain a new point set $\mathcal{P}^{G}_{j}$. Then, all points belonging to $\mathcal{P}^{G}_{j}$ are transferred to the intermediate plane $\mathbf{P}$:
	\begin{equation}	
			\textbf{p}^{\prime}_{j} = \mathbf{T}_{in} \cdot \textbf{p}_{G}=\mathbf{R}_{in} \cdot \textbf{p}_{G}+\mathbf{t}_{in}, \\
		\label{trans2}
	\end{equation}
	where $\textbf{p}^{\prime}_{j}$ is the point transmitted to the intermediate plane. $\mathbf{T}_{in} = [\mathbf{R}_{in}|\mathbf{t}_{in}]$ where $\mathbf{R}_{in} = \mathbf{I}_{3 \times 3}$ is an identity matrix and $\mathbf{t}_{in} = [1 \mathrm{m}, 0, 0]^{T} $ since the plane equation is $x=1 \mathrm{m}$. Define the point set on the intermediate plane as $\mathcal{P}^{\prime}_{j}$. Given that $\mathcal{P}^{\prime}_{j}$ is a point cloud with no occlusions, tag localization in it is straightforward using IFM \cite{IFM}. IFM returns the tag ID and the locations of the vertices labeled by index if the tag is found within the candidate OBB. Then, the locations of vertices can be transferred back to the original positions in the 3D map using the reverse processes of Eqs. (\ref{trans}-\ref{trans2}), and the poses of tags (from $\{\mathbf{G}\}$ to the tag coordinate system) are obtained by solving Singular Value Decomposition \cite{IFM}. The final tag localization result is presented in Fig. \ref{step5}. \par

 \noindent\textbf{Consideration of the rotation ambiguity of 3D OBB.} As shown in Fig. 11(a), a 3D OBB is symmetric, and thus, $\mathbf{T}_{OBB}$ could suffer from ambiguous rotation. This will result in a flipped image when the points are transferred to the intermediate plane. However, for fiducial tag patterns, if a pattern exists in the coding library, the flipped pattern will not \cite{ap3,aruco}. Thus, in practice, we only need to check both the raw intensity image and the flipped intensity image to resolve the rotation ambiguity.\par
  \noindent\textbf{Pose estimation accuracy boost brought by the intermediate plane.} Suppose there is no occlusion, and IFM also functions. Transferring the point cloud of a fiducial tag onto the intermediate plane physically reduces the distance between the point cluster and the LiDAR, resulting in a larger projection. This suggests that our intensity image has better imaging quality compared to IFM, which projects the raw point cloud directly onto the imaging plane. A better imaging quality is beneficial to pose estimation \cite{IFM}.
 % Define the point set composed of the detected 3D fiducials as $\mathcal{F}_{j}$. If the set $\mathcal{F}{j}$ exists, we can transmit the points in $\mathcal{F}{j}$ from the intermediate plane back to their original locations using the reverse processes of Eqs. (\ref{trans}-\ref{trans2}). 
 \par
	 \par
\begin{figure}[thpb]
		\centering
		\includegraphics[width=3.3in]{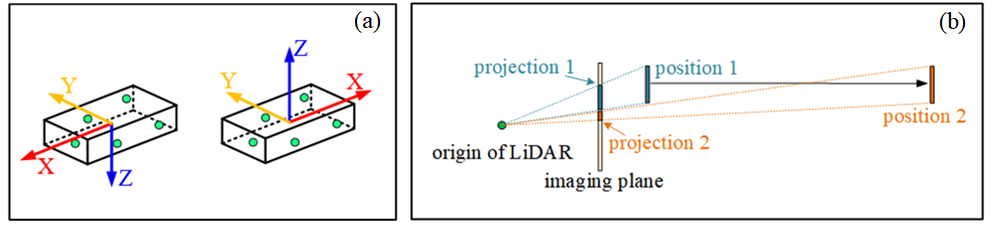}
		\caption{(a): An illustration of the rotation ambiguity issue of a 3D OBB. (b): An illustration of why the pose estimation improves with the intermediate plane.}
		\label{add}
	\end{figure} \par
 \vspace{-0.2cm}
\noindent\textbf{Remark 4.} 
We want to clarify that for cases where Eq. (\ref{pro}) is applicable, tag detection is the contribution of \cite{IFM}, rather than of this work. The contributions of this work lie in localizing fiducial tags on a 3D LiDAR map where Eq. (\ref{pro}) is not applicable. This is achieved through the proposed new pipeline for jointly analyzing a 3D map from intensity and geometry perspectives, introduced in sections \ref{ic}, \ref{clu}, \ref{clu2}, and \ref{ip}. \par
 \begin{table}[htbp]
\caption{Comparison with previous methods across different scenes.}
	\centering
	\resizebox{0.7\columnwidth}{!}{
 % {
	% \begin{center}
		\begin{tabular}{c|c|c|c|c}
			\hline\hline
				Scene &  LiDARTag \cite{lt} & A4 \cite{a4} & IFM \cite{IFM} & Ours  \\ \hline
    Fig. \ref{sub} & 0 / 2 & 0 / 2 & 1 / 2 & 2 / 2   \\ \hline
    Fig. \ref{mov} & 0 / 1 & 0 / 1 & 0 / 1 & 1 / 1   \\ \hline
    Fig. \ref{map2} & 0 / 3 & 0 / 3 & 0 / 3 & 2 / 3   \\
  
\hline  \hline
			
		\end{tabular}
	}
\label{tab1}
\end{table}

\begin{table*}[htbp]
\caption{Comparison of our approach, IFM \cite{IFM}, and AprilTag3 \cite{ap3} with respect to pose estimation accuracy.}
\begin{center}

\resizebox{1.5\columnwidth}{!}{

\begin{tabular}{c|c|c|c|c|c|c|c}
\hline\hline
%&\multicolumn{3}{|c|}{\textbf{Table Column Head}} \\
 Distance& Method & X error (m) & Y error (m) &Z error (m)&  Roll error (deg)&  Pitch error (deg) & Yaw error (deg) \\
\cline{1-8} 
\multirow{3}{*}{ 2m} & AprilTag3 \cite{ap3} & 0.016 & 0.034 &0.016& 0.807&  3.166 &  1.244 \\ \cline{2-8} 
& IFM \cite{IFM} & 0.003 & 0.006 &0.017& 0.423&  0.457 &  0.399 \\ \cline{2-8} 
& \textbf{Ours} &\textbf{ 0.002} &\textbf{ 0.005}  &\textbf{0.011}&\textbf{ 0.315}&  \textbf{0.305} & \textbf{ 0.391} \\ \cline{1-8}

\multirow{3}{*}{ 3m} & AprilTag3 \cite{ap3} & 0.058 & 0.124 &0.044& 1.369&  7.963 & 2.904 \\ \cline{2-8} 
& IFM \cite{IFM} & 0.026  & 0.021 &0.093& 0.930&  1.182 &  0.859 \\ \cline{2-8} 
& \textbf{Ours} &\textbf{ 0.006} &\textbf{ 0.009} &\textbf{0.015}&\textbf{ 0.343}& \textbf{ 0.322} & \textbf{ 0.455} \\ \cline{1-8}

\multirow{3}{*}{ 4m} & AprilTag3 \cite{ap3} & 0.072 & 0.407 &0.233& 1.433&  9.292 &  13.343 \\ \cline{2-8} 
& IFM \cite{IFM} & 0.024 & 0.024 &0.107& 0.342&  2.020 &  1.111 \\ \cline{2-8} 
& \textbf{Ours} &\textbf{ 0.008} & \textbf{0.014} &\textbf{0.016}&\textbf{ 0.302}&  \textbf{0.389} &\textbf{ 0.478} \\ \cline{1-8}
\hline\hline
\end{tabular}
\label{tab2}
}
\end{center}
\end{table*}
 \par
\vspace{-0.1cm}
\section{Experimental Results\label{exp}}
 %    	\begin{figure*}
	% 	\centering
	% 	\includegraphics[width=17.6cm]{process.png}
	% 	\caption{  Visual results of the main steps. The raw point cloud is shown in Fig. \ref{sub}. (a): The point cloud after downsampling using 3D intensity gradients. (b): Clustering of the downsampling result. Each cluster is labeled with an OBB. (c): Preserved OBBs according to the proposed criterion. (d): Points extracted from the raw point cloud and the points transmitted to the intermediate plane. (e): LiDAR fiducial markers detection result. The detected vertices are rendered in red.}
	% 	\label{process}
	% \end{figure*}\par
% In this section, we present experimental results comparing the proposed approach to existing methods, including IFM \cite{IFM}, LiDARTag \cite{lt}, and A4LiDARTag \cite{a4}. 
\subsection{Qualitative Evaluation} \label{general}

\begin{figure}[thpb]
		\centering
		\includegraphics[width=3.3in]{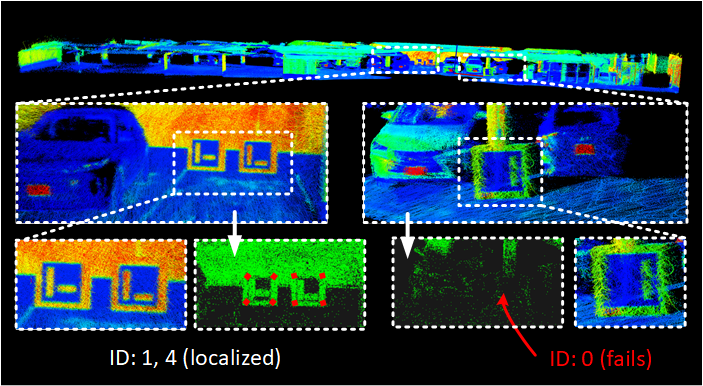}
		\caption{  This is an indoor parking lot scene with three ArUco tags \cite{aruco} placed in it. The LiDAR scans the parking lot from right to left (scans tag 0 before tags 1 and 4). However, it moves faster when scanning tag 0, leading to a sparser point cloud and more severe motion blur.}
		\label{map2}
	\end{figure}

We first compare the existing methods \cite{IFM,lt,a4} with our approach in terms of tag detection rate. In addition to the map of campus (Fig. \ref{mov}) and the synthesis scene (Fig. \ref{sub}), we also conducted experiments in an indoor parking lot scene, as shown in Fig. \ref{map2}. The 3D maps presented in Fig. \ref{mov} and Fig. \ref{map2} are constructed using the Livox MID-40 LiDAR \cite{loam} with the \textit{livox mapping} approach \cite{sdk}.
The result is presented in Table \ref{tab1}. 
%
% The visual results after completing the major steps of the proposed approach on the 3D LiDAR map shown in Fig. \ref{mov} are presented in Fig. \ref{map}. 
%
As seen, the proposed approach has the highest tag detection rate.

 % \begin{figure}[thpb]
	% 	\centering
	% 	\includegraphics[width=3.3in]{map}
	% 	\caption{  An illustration of the tag localization process on the 3D LiDAR map shown in Fig. \ref{mov}. (a): After sections \ref{ic} and \ref{clu}. (b): After section \ref{clu2}. As seen, all the clusters but the one belonging to the tag are filtered out. (c): The zoomed view of the preserved OBB. (d): A zoomed view of the detected 3D fiducials, which are rendered in red.}
	% 	\label{map}
	% \end{figure} 

 	\begin{figure}[thpb]
		\centering
		\includegraphics[width=3.0in]{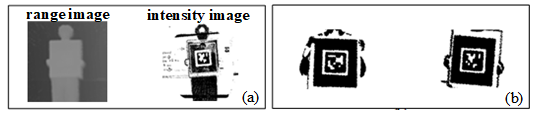}
		\caption{ (a): The 2D range and intensity images after projecting the 3D point cloud shown in Fig. \ref{sub} onto the 2D image plane. (b): The intensity images obtained through our approach from the point cloud shown in Fig. \ref{sub}. }
		\label{inten}
	\end{figure} \par

We take the simple scene depicted in Fig. \ref{sub} as an example to analyze the reasons for the detection failures of the existing methods. LiDARTag \cite{lt} and A4LiDARTag \cite{a4} require range information to detect fiducials. However, as shown in the range image in Fig. \ref{inten}(a), it lacks useful information. IFM \cite{IFM} only detects one tag since only one tag is preserved on the intensity image (see Fig. \ref{inten}(a)). By contrast, our approach generates an intensity image for each point cluster that could contain a fiducial tag (see Fig. \ref{inten}(b)). \par
\noindent\textbf{Limitations.} As seen in Fig. \ref{map2}, the proposed method only detects two tags with good scan quality out of the three tags in the scene. Namely, poor scan quality (sparse or motion-blurred point cloud) will lead to localization failure. Thus, in practice, if we use a low-cost non-repetitive scanning LiDAR, such as the Livox MID-40, we have to scan the scene slowly to obtain a dense map and avoid motion blur.
 \begin{figure}[thpb]
		\centering
		\includegraphics[width=1.7in]{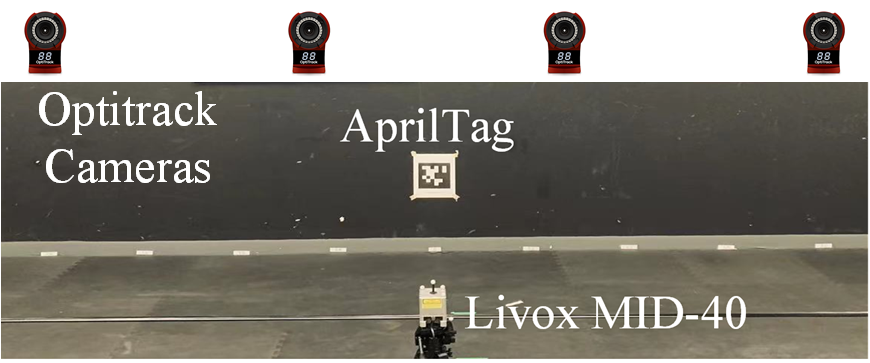}
		\caption{ The experimental setup for quantitative evaluation. A Letter size
AprilTag is put on the wall. The MoCap system provides the ground truth.}
		\label{setup}
	\end{figure} 
 \vspace{-0.2cm}
\subsection{Quantitative Evaluation} \label{quan}
In this section, we conduct experiments introduced in Fig. \ref{setup} to demonstrate that the proposed method has better pose estimation accuracy. Since we want to compare our method with other approaches in terms of pose estimation accuracy, it is necessary to use this occlusion-free scene because other approaches are not applicable to point clouds with occlusion. We test the methods under different distances from the LiDAR to the fiducial tag. Unfortunately, LiDARTag \cite{lt} and A4LiDARTag \cite{a4} fail to detect the tag, and thus, we compare the proposed approach with IFM \cite{IFM} and the widely-used AprilTag3 \cite{ap3}. Note that when testing AprilTag3, the adopted sensor is a camera. As shown in Table \ref{tab2}, both the accuracy of IFM and AprilTag3 degrades as the distance from the sensor to the marker increases while the accuracy degradation of IFM is less severe compared to AprilTag3. The accuracy of the proposed approach is slightly better than IFM at a distance of 2m. However, unlike IFM, our approach maintains a decent accuracy when the distance increases. The reason for the accuracy boost has been analyzed in the subsection \textbf{Pose estimation accuracy boost brought by the intermediate plane} of section \ref{ip}. \par
%  \begin{figure}[thpb]
% 		\centering
% 		\includegraphics[width=3.3in]{projs}
% 		\caption{ An illustration of the negative correlation between the projection size and the distance. It is straightforward to prove the negative correlation using the similarity of triangles.}
% 		\label{projs}
% 	\end{figure}
% \begin{figure}[thpb]
% 		\centering
% 		\includegraphics[width=2.5in]{pics}
% 		\caption{ The visual comparison of the intensity images obtained by our method and IFM \cite{IFM} from the same point cloud (shown on the right side) when the distance from the marker to the LiDAR is 3m.}
% 		\label{pics}
% 	\end{figure} \par
% The analysis of the result is given as follows. Suppose that both IFM \cite{IFM} and our approach are employed to process the same point cloud where the distance from the fiducial marker to the LiDAR is 3m. IFM \cite{IFM} projects the entire raw point cloud to the image plane directly without any analysis. By contrast, our approach first transforms the points potentially corresponding to the marker to the intermediate plane and then projects them to the image plane. Based on the negative correlation between the projection size and distance, as illustrated in Fig. \ref{projs}, our method results in a larger marker projection on the image plane than IFM \cite{IFM}. The visual comparison is given in Fig. \ref{pics}. Apparently, using our approach, the intensity image has fewer points overlapping and thus retains more information. This is favorable for pose estimation as it leads to more accurate observations.
\section{Conclusions and Future Work\label{con}}
In this work, we develop a novel method to localize fiducial tags on a 3D LiDAR prior map. Despite the existence of previous LiDAR fiducial tag detection methods \cite{IFM,lt,a4} for the LiDAR sensor, our method is the first that applies to a 3D LiDAR map. The proposed pipeline for jointly analyzing a point cloud from the intensity and geometry can directly localize fiducial tags in a 3D map, returning tag poses labeled by ID number and vertice positions labeled by vertice index. In addition, the proposed intermediate-plane-based method boosts the pose estimation accuracy by generating intensity images with better quality than IFM \cite{IFM}. With accurately localized fiducial tags on a 3D LiDAR map, an unmanned system equipped with a LiDAR can obtain positioning information relative to the map by scanning the tags, thus enabling navigation unaffected by changes in ambient light. Since we lack the equipment to obtain ground truth outdoor positioning, we only use MoCap indoors to evaluate the accuracy of the proposed method. In the future, we plan to quantitatively evaluate the performance of our method outdoors and integrate it with LiDAR-based relocalization and navigation.
 
% 

%%%%%%%%%%%%%%%%%%%%%%%%%%%%%%%%%%%%%%%%%%%%%%%%%%%%%%%%%%%%%%%%%%%%%%%%%%%%%%%%
	%\section*{APPENDIX}
	% \section*{ACKNOWLEDGMENT}
	% The authors would like to thank Yicong Fu, Guile Wu, Han Wang, Brian Lynch, Yuan Ren, Shuo Zhang, Marc Savoie, and Shiyuan Jia for insightful discussions. 
	%
	%
	%
	%%%%%%%%%%%%%%%%%%%%%%%%%%%%%%%%%%%%%%%%%%%%%%%%%%%%%%%%%%%%%%%%%%%%%%%%%%%%%%%%%
	%
	%References are important to the reader; afore, each citation must be complete and correct. If at all possible, references should be commonly available publications.

	\bibliographystyle{IEEEtran} 
	\bibliography{reference} 

\end{document}